\documentclass[aps,pre,reprint,amsmath,amssymb]{revtex4-2}
\usepackage{txfonts}
\usepackage{algorithmic}
\usepackage{algorithm}
\usepackage{graphicx}
\usepackage{bm}
\usepackage{url}
\usepackage{mathrsfs}
\usepackage{listings}
\usepackage{color}

\begin{document}

\title{Aspects of importance sampling in parameter selection for neural networks using ridgelet transform}

\author{Hikaru Homma and Jun Ohkubo}

\affiliation{Graduate School of Science and Engineering, Saitama University, Sakura, Saitama 338--8570, Japan}

\begin{abstract}
The choice of parameters in neural networks is crucial in the performance, and an oracle distribution derived from the ridgelet transform enables us to obtain suitable initial parameters. In other words, the distribution of parameters is connected to the integral representation of target functions. The oracle distribution allows us to avoid the conventional backpropagation learning process; only a linear regression is enough to construct the neural network in simple cases. This study provides a new look at the oracle distributions and ridgelet transforms, i.e., an aspect of importance sampling. In addition, we propose extensions of the parameter sampling methods. We demonstrate the aspect of importance sampling and the proposed sampling algorithms via one-dimensional and high-dimensional examples; the results imply that the magnitude of weight parameters could be more crucial than the intercept parameters.
\end{abstract}

\maketitle

\section{Introduction}
\label{sec_introduction}

Recently, the fields of machine learning and artificial intelligence have developed rapidly. In particular, deep neural networks are available for various situations. Here, `deep' means that the neural network has many intermediate layers; the greater number of layers yields higher learning ability. However, a `shallow' neural network is sometimes sufficient to represent enough learning ability \cite{Barron1993,Murata1996}. Since the deep structure prevents our understanding of neural networks, studies on shallow neural networks from a physics perspective would be beneficial. In Ref.~\cite{Sonoda2014}, simple sampling-based learning methods for the shallow neural networks were proposed. Probability distributions for weight and intercept parameters, so-called oracle distributions, are constructed via the ridgelet transform; the oracle distributions show the usefulness of the parameters for neural networks. It is also possible to employ conventional learning methods based on backpropagation, in which we can see the role of ridgelet transform and oracle distributions as adequate initialization for neural networks. Of course, there are previous works on initializing neural network parameters; for example, see Refs.~\cite{Denoeux1993,Freitas2000,LeCun2012}. Although it would be typical to use random initialization in practice, the initialization based on the ridgelet transform shows preferable behavior in the test accuracy \cite{Sonoda2014}.

The shallow neural networks have attracted attention from other perspectives; extreme learning machines (ELMs) are training algorithms for a feedforward neural network with only one hidden layer \cite{Huang2004,Huang2006}. In the ELMs, the parameters on the hidden nodes are randomly chosen. The learning process is only on the output layer via a simple linear regression, which is faster and more efficient than conventional methods based on backpropagation. As for recent developments in the ELMs, see a review paper \cite{Wang2022}. It is easy to imagine that one could employ the ridgelet transforms and the oracle distributions instead of the random choices of the hidden parameters in the ELMs.

In the present paper, we provide a new perspective on the sampling method based on the ridgelet transforms. As discussed in Ref.~\cite{Sonoda2014}, samplings from exact oracle distributions are difficult because of the high dimensionality. Hence, some approximations were employed, as reviewed later. However, there are some heuristics in the approximated sampling method. To propose new sampling methods, we first provide an insight into the oracle distribution from an aspect of importance sampling. The importance sampling is one of the essential tools in physics; see, for example, Ref.~\cite{Robert_book}. As we see later, the importance sampling is insufficient to make a practical algorithm, and therefore, we propose two additional sampling methods. Numerical experiments on one-dimensional and higher-dimensional cases provide insight into which parts of the network parameters are crucial.

The structure of this paper is as follows. In Sect.~\ref{sec_previous}, we briefly review the previously proposed methods. Section~\ref{sec_importance_sampling} provides discussions on the oracle distribution from the aspect of importance sampling. Section~\ref{sec_proposal} presents extensions of the practical sampling method in the previous study \cite{Sonoda2014}. Section~\ref{sec_conclusion} denotes concluding remarks.

\section{Previous Work on Ridgelet Transforms}
\label{sec_previous}

In this section, we briefly review the previous study of oracle distributions \cite{Sonoda2014}. For details, see Refs.~\cite{Sonoda2014} and \cite{Sonoda2017}. Although complex functions should be included in general discussions on ridgelet transforms, we avoid the usage of complex functions as far as possible for practicality and simplicity.

\subsection{Neural networks and ridgelet transforms}

Consider a neural network with only one hidden layer, which approximates some function $f:\mathbb{R}^m \rightarrow \mathbb{R}$. Let $\eta:\mathbb{R}\rightarrow\mathbb{R}$ be an activation function. Then, the neural network is expressed as
\begin{align}
g_J(\bm{x}) = \sum_{j=1}^{J} c_{j}\eta (\bm{a}_{j} \cdot \bm{x} - b_{j}) + c_0,
\label{eq_NN}
\end{align}
where $J$ is the number of hidden nodes, $\{\bm{a}_j\}$ and $\{b_j\}$ are weight and intercept parameters in the hidden layer, respectively, $\{c_j\}$ are output parameters, and $c_0$ corresponds to the intercept term on the output layer.

In Ref.~\cite{Sonoda2014}, the neural network of Eq.~\eqref{eq_NN} is connected to an integral representation via a ridgelet transform. The ridgelet transform $\mathscr{R}_{\psi}f$ of $f: \mathbb{R}^m \rightarrow \mathbb{R}$ with respect to $\psi: \mathbb{R} \rightarrow \mathbb{R}$ is formally given by
\begin{align}
\mathscr{R}_{\psi}f(\bm{a},b) =\int_{\mathbb{R}^{m}}f(\bm{x}) \psi (\bm{a} \cdot \bm{x} - b) d\bm{x},
\label{eq_ridgelet}
\end{align}
where $(\bm{a},b) \in \mathbb{R}^{m+1}$. In general, it is possible to consider the ridgelet transform with a multiplication of $ \|\bm{a}\|^s$ to the the integrand in Eq.~\eqref{eq_ridgelet} \cite{Sonoda2017}. However, the transformation with $s=0$ is suitable for the Euclidean formulation and employed in the previous work \cite{Sonoda2014}. Hence, we employ the definition of Eq.~\eqref{eq_ridgelet} in this study. 

The dual ridgelet transform $\mathscr{R}_{\eta}^{\dagger} T$ of $T: \mathbb{R}^{m+1} \rightarrow \mathbb{R}$ with respect to $\eta: \mathbb{R} \rightarrow \mathbb{R}$ is given by
\begin{align}
\mathscr{R}_{\eta}^{\dagger} T(\bm{x}) =& \int_{\mathbb{R}^{m+1}} T(\bm{a},b)\eta (\bm{a} \cdot \bm{x} - b)d\bm{a}db.
\label{eq_dual_ridgelet}
\end{align}
Two functions $\psi$ and $\eta$ are said to be admissible when the following quantity is finite and not zero:
\begin{align}
K_{\psi,\eta} = (2\pi)^{m-1}\int_{-\infty}^{\infty}\frac{\overline{\hat{\psi}(\bm{\xi})}\hat{\eta}(\bm{\xi})}{|\bm{\xi}|^{m}}d\bm{\xi},
\end{align}
where $\hat{\cdot}$ denotes the Fourier transform, and $\overline{\,\cdot\,}$ indicates the complex conjugate. In the following discussion, we suppose that $\psi, \eta$, and $f$ belong to certain adequate classes \cite{Sonoda2017}, and $\psi$ and $\eta$ are admissible. Then, the following reconstruction formula holds:
\begin{align}
\mathscr{R}_{\eta}^{\dagger}\mathscr{R}_{\psi} f &= K_{\psi,\eta}f.
\label{eq_reconstruction_formula}
\end{align}
Note that Eq.~\eqref{eq_reconstruction_formula} leads to
\begin{align}
f(\bm{x}) &=
\frac{1}{K_{\psi,\eta}}  \mathscr{R}_{\eta}^{\dagger}\mathscr{R}_{\psi}f(\bm{x}) \nonumber \\
&= \frac{1}{K_{\psi,\eta}} \int_{\mathbb{R}^{m+1}} \mathscr{R}_{\psi}f(\bm{a},b) \eta (\bm{a} \cdot \bm{x} - b) d\bm{a}db \nonumber \\
&= \frac{1}{K_{\psi,\eta}}  \int_{\mathbb{R}^{m+1}} \widetilde{c}(\bm{a},b) \eta (\bm{a} \cdot \bm{x} - b) \mu(\bm{a},b) d\bm{a}db,
\label{eq_basis_reconstruction_integral}
\end{align}
where we introduce a probability measure $\mu(\bm{a},b)$ and a function $\widetilde{c}(\bm{a},b)$ via
\begin{align}
\mathscr{R}_{\psi}f(\bm{a},b) = \widetilde{c}(\bm{a},b) \mu(\bm{a},b).
\end{align}
The probability measure $\mu(\bm{a},b)$ is called an oracle distribution. Assume $\{(\bm{a}_j, b_j)\}_{j=1}^J$ is sampled from the measure $\mu(\bm{a},b)$. Then, the approximation of the integral in Eq.~\eqref{eq_basis_reconstruction_integral} with a finite summation yields
\begin{align}
f(\bm{x}) &\simeq
 \frac{1}{K_{\psi,\eta}}  \frac{1}{J} \sum_{j=1}^J \widetilde{c}(\bm{a}_j,b_j) \eta (\bm{a}_j \cdot \bm{x} - b_j) + c_0,
\label{eq_basis_reconstruction}
\end{align}
where a constant $c_0$ complements the discrepancy due to the approximation of the integral in Eq.~\eqref{eq_basis_reconstruction_integral}. We immediately see that Eq.~\eqref{eq_basis_reconstruction} corresponds to the neural network of Eq.~\eqref{eq_NN} with the replacement of $\widetilde{c}(\bm{a}_j,b_j)/(K_{\psi,\eta} J)$ with $c_j$. That is, $g_{J}(\bm{x})$ is regarded as a discrete approximation of $f(\bm{x})$, and hence, Eq.~\eqref{eq_basis_reconstruction_integral} is called an integral representation of neural networks.

\subsection{Sampling of hidden parameters}

Sonoda and Murata gave practical algorithms for the sampling of $(\bm{a},b)$ from $\mu(\bm{a},b)$ \cite{Sonoda2014}. We briefly review them.

It may be possible to directly compute $\mathscr{R}_{\psi}f(\bm{a},b)$ using an explicit activation function $\eta$ and the function $\psi$ for the ridgelet transformation. However, direct computation is not feasible for high-dimensional cases. Hence, the Monte Carlo approximation is employed here. Furthermore, some approximations were introduced in Ref.~\cite{Sonoda2014} since a naive acceptance-rejection sampling method is inefficient and unstable in high-dimensional cases.

Consider a dataset $\{(\bm{x}_n,y_n)\}_{n=1}^N$. Then, the integral in Eq.~\eqref{eq_ridgelet} is approximated as follows:
\begin{align}
\mathscr{R}_{\psi}f(\bm{a},b)
&=  \int_{\mathbb{R}^m} f(\bm{x}) \psi (\bm{a} \cdot \bm{x} - b)d\bm{x} \nonumber \\
&\propto
\sum_{n=1}^{N} y_{n}  \psi({\bm{a}} \cdot \bm{x}_{n} - b),
\label{eq_ridgelet_approximation}
\end{align}
where we assume that $\{\bm{x}_n\}$ are sampled uniformly.

Next, we consider the following upper bound for $\mu(\bm{a},b)$:
\begin{align}
\mu(\bm{a},b) 
&\propto \left\lvert \mathscr{R}_{\psi}f(\bm{a},b)\right\rvert
\propto \left\lvert \sum_{n=1}^{N} y_{n} \psi({\bm{a}} \cdot \bm{x}_{n} - b) \right\rvert \nonumber \\
&\leq \sum_{n=1}^{N} \left\lvert y_{n} \right\rvert
\left\lvert \psi(\bm{a} \cdot \bm{x}_{n} - b) \right\rvert 
\propto\sum_{n=1}^{N}\omega_{n}\mu_{n}(\bm{a},b),
\label{eq_mixture_approximation}
\end{align}
where $\omega_n \propto \lvert y_n \rvert$ and $\mu_n(\bm{a},b)\propto|\psi(\bm{a}\cdot \bm{x}_n-b)|$. The distribution given by the final expression in Eq.~\eqref{eq_mixture_approximation} is called the mixture distribution \cite{Sonoda2014}. Hence, the following two-step sampling algorithm is available: choose an index $n$ according to the mixing probability $\omega_n$; draw a sample $(\bm{a},b)$ from $\mu_{n}(\bm{a},b)$. 

In Ref.~\cite{Sonoda2014}, a further extension of the sampling algorithm was proposed, i.e., a sampling from a mixture annealed distribution. In the extended algorithm, $\mu_{n}(\bm{a},b)$ is approximated or annealed by a beta distribution $\mathrm{Beta}(\alpha,\beta)$, which has almost all mass around both ends of its domain and nearly no mass in the center. Note that $\mathrm{Beta}(100,3)$ was employed in Ref.~\cite{Sonoda2014}. In the extended algorithm, we first generate $z$ from the beta distribution. Then, $(\bm{a},b)$ is generated under a restriction $z=\bm{a}\cdot \bm{x}_n-b$. Since various $(\bm{a},b)$ are fulfilling the restriction, two additional assumptions were introduced \cite{Sonoda2014}:
\begin{itemize}
\item $\bm{a}$ is parallel to a given $\bm{x}_n$.
\item $|\bm{a}|^{-1}$ has a similar scale to distances between two input vectors. (Practically, we randomly select two inputs $\bm{x}_n$ and $\bm{x}_m$, and set $|\bm{a}| = |\bm{x}_n-\bm{x}_m|^{-1}$.)
\end{itemize}

Finally, the algorithm~\ref{alg1} is derived.

\begin{algorithm}[H]
\caption{Basic sampling algorithm}
\label{alg1}
\begin{algorithmic}[1]
\STATE Choose $n$ and $m$ according to the probability $\{\omega_{n}\}$.
\STATE Draw $\zeta\sim \mathrm{Beta}(\alpha,\beta)$ and $\gamma\sim \mathrm{Bernoulli}(\gamma;p=0.5)$
\STATE $z\leftarrow(-1)^{\gamma}\zeta$
\STATE $1/a \leftarrow|\bm{x}_{n}-\bm{x}_{m}|$
\STATE $\bm{a}\leftarrow a \bm{x}_{n}/|\bm{x}_{n}|$
\STATE $b\leftarrow \bm{a}\cdot \bm{x}_{n}-z$
\STATE Return $(\bm{a},b)$
\end{algorithmic}
\end{algorithm}

\subsection{Sampling of output parameters}

After the sampling with Algorithm~\ref{alg1} yields $\{(\bm{a}_j,b)\}_{j=1}^J$, the output parameters $\{c_j\}_{j=1}^J$ should be determined. Note that there is no nonlinear function on the output node. Hence, the least squares method with the dataset $\{(\bm{x}_n,y_n)\}_{n=1}^N$ leads to $\{c_j\}_{j=1}^J$ immediately. Although a simple linear regression was employed in Ref.~\cite{Sonoda2014}, ridge or least absolute shrinkage and selection operator (lasso) regressions are also available.

\subsection{Numerical examples}
\label{subsec_numerical_example_TSC}

\begin{figure}[bt]
\begin{center}
\includegraphics[width=65mm]{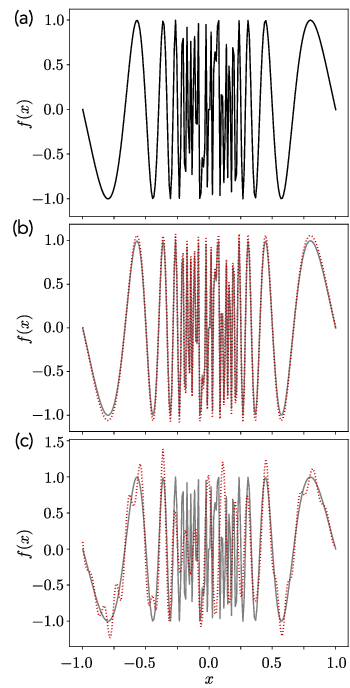}
\caption{(Color online) (a) The topologist's sine curve (TSC). (b) $f(\bm{x})$ obtained by Eq.~\eqref{eq_basis_reconstruction_integral}. (c) $g_J(\bm{x})$ obtained by Algorithm~\ref{alg1} and the ridge regression. The solid line corresponds to the original curve, which has numerical instability around $x=0$. The dotted curves in (b) and (c) correspond to the approximated ones.}
\label{fig_previous_methods}
\end{center}
\end{figure}

As a demonstration, we here give examples computed by the methods described above. The objective function is a topologist's sine curve (TSC) used in Ref.~\cite{Sonoda2014},
\begin{align}
f(x)=\mathrm{sin}\frac{2\pi}{x},
\label{eq_TSC}
\end{align}
defined on $x \in [-1,1]$ with $f(0)=0$. Figure~\ref{fig_previous_methods}(a) shows the original curve of the TSC, which is a complicated curve whose spatial frequency gets quite high around $x=0$. There are $200$ points sampled from $[-1,1]$ in an equidistant manner in Fig.~\ref{fig_previous_methods}(a). 

Here, we use the following functions in the ridgelet transform \cite{Sonoda2017}:
\begin{align}
\eta(x)&=\exp{\left(-\frac{x^2}{2}\right)},\\
\psi(x)&=\frac{1}{\pi^{2}}\left(2x(x^{2}-3)F\left(\frac{x}{\sqrt{2}}\right)-\sqrt{2}(x^{2}-2)\right),
\end{align}
where $F(x)=e^{-x^{2}}\int_{0}^{x}e^{z^{2}}dz$ is known as the Dawson's integral. They meet the admissibility condition and $K_{\psi,\eta} = 1$.

Since this example has only one-dimensional input space, a discrete lattice is available to evaluate the integral in Eq.~\eqref{eq_basis_reconstruction_integral}. Here, we assume the integral domain with $x \in [-1,1]$ and the discrete lattice with $\Delta x = 0.01$ for Eq.~\eqref{eq_ridgelet}; for Eq.~\eqref{eq_dual_ridgelet}, we employ $a \in [-300,300]$, $\Delta a = 0.1$, $b \in [-300,300]$, and $\Delta b = 0.1$. Figure~\ref{fig_previous_methods}(b) shows the result via the discrete approximation of the integral. We see that the original TSC curve is adequately recovered except for some errors which stem from the discrete approximations.

For Algorithm~\ref{alg1}, we set $J=300$, and the parameters for the Beta distribution are $\alpha = 50$ and $\beta = 3$, i.e., $\mathrm{Beta}(50,3)$. The training dataset $\{(x_n, y_n)\}$ is generated with the function values $y_n = f(x_n)$ of $200$ points $\{x_n\}$ extracted at equal intervals from $[-1, 1]$. The output parameters $\{c_j\}$ are determined by the ridge regression with the regularization parameter of $0.01$. Figure~\ref{fig_previous_methods}(c) shows the result; the sampling method recovers a rough shape of the TSC despite the coarse approximation.

\section{Aspect as Importance Sampling}
\label{sec_importance_sampling}

In this section, we provide a new viewpoint for the previous work in Ref.~\cite{Sonoda2014} from the aspect of the importance sampling.

\subsection{Usage of sampling distribution}

We introduce probability density functions $\rho^{(1)}(\bm{x})$ and  $\rho^{(2)}(\bm{a},b)$. Then, Eq.~\eqref{eq_ridgelet} is rewritten as follows:
\begin{align}
\mathscr{R}_{\psi}f(\bm{a},b)
&=\int_{\mathbb{R}^{m}}^{} f(\bm{x})\psi (\bm{a}\cdot \bm{x}-b) d\bm{x} \nonumber\\
&=\int_{\mathbb{R}^{m}}^{} \rho^{(1)}(\bm{x}) 
\frac{1}{\rho^{(1)}(\bm{x})} f(\bm{x})\psi (\bm{a}\cdot \bm{x}-b) d\bm{x} \nonumber \\
&\simeq  \frac{1}{N} \sum_{n=1}^{N} \frac{1}{\rho^{(1)}(\bm{x}_{n})}
y_n \psi (\bm{a}\cdot \bm{x}_n-b),
\label{eq_IS_1}
\end{align}
where $\{\bm{x}_n\}$ are drawn from $\rho^{(1)}(\bm{x})$. Equation~\eqref{eq_basis_reconstruction_integral} is also rewritten as follows:
\begin{align}
f(\bm{x})
&= \frac{1}{K_{\psi,\eta}} \int_{\mathbb{R}^{m+1}} \mathscr{R}_{\psi}f(\bm{a},b) \eta (\bm{a} \cdot \bm{x} - b) d\bm{a}db \nonumber \\
&= \frac{1}{K_{\psi,\eta}} \int_{\mathbb{R}^{m+1}} \rho^{(2)}(\bm{a},b) \frac{\mathscr{R}_{\psi}f(\bm{a},b)}{\rho^{(2)}(\bm{a},b)} \eta (\bm{a}\cdot \bm{x}-b) d\bm{a}db \nonumber \\
&\simeq \frac{1}{K_{\psi,\eta}} \frac{1}{J} \sum_{j=1}^{J} \frac{\mathscr{R}_{\psi}(\bm{a}_{j},b_{j})}{\rho^{(2)}(\bm{a}_{j},b_{j})} \eta (\bm{a}_j\cdot \bm{x}-b_j) + c_0,
\label{eq_IS_2}
\end{align}
where $\{(\bm{a}_{j},b_{j})\}$ are drawn from $\rho^{(2)}(\bm{a},b)$. Note that we assume the conditions $\rho^{(1)}(\bm{x}_n) \neq 0$ for all $n$ and $\rho^{(2)}(\bm{a}_{j},b_{j}) \neq 0$ for all $j$, for simplicity. Then, defining
\begin{align}
c_{j} &\equiv 
\frac{1}{K_{\psi,\eta}}\frac{1}{J}\frac{\mathscr{R}_{\psi}(\bm{a}_{j},b_{j})}{\rho^{(2)}(\bm{a}_{j},b_{j})},
\label{eq_IS_3}
\end{align}
it is possible to determine the output parameters $\{c_j\}$ without using the linear regression except for the intercept term $c_0$. 

We can see the above discussions as an application of importance sampling, in which we sample important parts of the integral evaluations. In principle, it is possible to use arbitrary density distributions for $\rho^{(1)}(\bm{x})$ and $\rho^{(2)}(\bm{a},b)$. For example, one could estimate $\rho^{(1)}(\bm{x})$ from the dataset $\{(\bm{x}_n,y_n)\}_{n=1}^N$ with the aid of a kernel density estimation. By contrast, there is no prior knowledge about $\rho^{(2)}(\bm{a}_{j},b_{j})$. Hence, we must use an assumed probability density. We next demonstrate a numerical example of the importance sampling aspect.

\subsection{Numerical experiments}

We demonstrate the above importance sampling aspect with a sin curve and the TSC example; for the TSC, the settings are the same as in Sect.~\ref{subsec_numerical_example_TSC}. Since the dataset $\{x_n\}$ is sampled from $[-1,1]$ in an equidistant manner, we set $\rho^{(1)}(x) = 1/2$ for $x \in [-1,1]$ and $N=200$. As described above, there is no prior knowledge about $\rho^{(2)}(a,b)$, and then we employ a bivariate normal distribution $\mathcal{N}(\bm{0}, \mathrm{diag}(100, 100))$. The number of hidden parameters is $J=300$, and we neglect the intercept term $c_0$.

\begin{figure}[bt]
\begin{center}
\includegraphics[width=65mm]{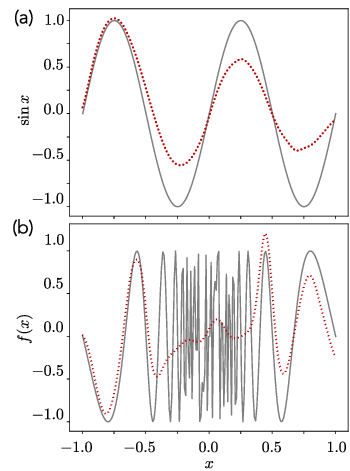}
\caption{(Color online) Function shapes obtained by Eqs.~\eqref{eq_IS_1}, \eqref{eq_IS_2}, and \eqref{eq_IS_3}. (a) and (b) correspond to $\sin (x)$ and $f(\bm{x})$, respectively. The solid and dotted curves correspond to the original functions and the obtained ones, respectively.}
\label{fig_is}
\end{center}
\end{figure}

Figure~\ref{fig_is} shows the numerical results. Note that the linear regression procedure is unnecessary to depict the dotted curve; only the sampling procedures are enough. Although the approximated functions do not yield the function shapes perfectly, the sampling procedures grasp the rough characteristics of the functions. We performed several numerical experiments using other types of $\rho^{(2)}(\bm{a},b)$, such as uniform distributions, and there was no significant difference in the obtained results.

These numerical experiments indicate that it is generally difficult to approximate the integrals of Eqs.~\eqref{eq_IS_1} and \eqref{eq_IS_2} with the importance sampling because there is no prior knowledge of density functions. Hence, the aid of the dataset and the regression procedure will help to improve the results, which supports the validity of the proposal in the previous study \cite{Sonoda2014}. However, the aspect of importance sampling enables us to extend Algorithm~\ref{alg1}; arbitrary density functions would be available to sample $(\bm{a},b)$. In the following section, we introduce two modified sampling algorithms different from Algorithm~\ref{alg1}.

\section{Proposal of Improved Sampling Methods}
\label{sec_proposal}

\subsection{Proposal 1: Sampling of the magnitude of $\bm{a}$}

On step 4 in Algorithm~\ref{alg1}, the magnitude of $\bm{a}$ is determined by the inverse of the distance between two inputs, $|\bm{x}_n-\bm{x}_m|$. We simply change it with an arbitrary density function. Of course, the density function with crucial contributions is desirable. We tried to use some density functions, including uniform and normal ones, and most of them worked. The normal distribution works well among them, and we propose Algorithm~\ref{alg2}. After sampling the hidden parameters $(\bm{a},b)$ by Algorithm~\ref{alg2}, the output parameters $\{c_j\}$ are determined by the ridge regression.

\begin{algorithm}[H]
\caption{Modified sampling algorithm I (Sampling of $\lvert\bm{a}\rvert$)}
\label{alg2}
\begin{algorithmic}[1]
\STATE Choose $n$ according to the probability $\{\omega_{n}\}$.
\STATE Draw $\zeta\sim \mathrm{Beta}(\alpha,\beta)$ and $\gamma\sim \mathrm{Bernoulli}(\gamma;p=0.5)$
\STATE $z\leftarrow(-1)^{\gamma}\zeta$
\STATE Draw $\kappa \sim\mathcal{N}(0,\delta)$
\STATE $\bm{a}\leftarrow \kappa\bm{x}_{n}/|\bm{x}_{n}|$
\STATE $b\leftarrow \bm{a}\cdot \bm{x}_{n}-z$
\STATE Return $(\bm{a},b)$
\end{algorithmic}
\end{algorithm}

\subsection{Proposal 2: Sampling of $b$}

Secondly, we propose an algorithm in which the parameter $b$ is sampled from an arbitrary density function. The assumption that $\bm{a}$ and $\bm{x}_n$ are parallel yields
\begin{align}
\bm{a} &= r\bm{x}_{n},
\end{align}
where $r$ is a constant. Then, the restriction $z = \bm{a} \cdot \bm{x}_n-b$ leads to
\begin{align}
\bm{a} \cdot \bm{x}_{n} = r\bm{x}_{n}\cdot \bm{x}_{n} &=b+z.
\end{align}
Hence, we have $r = (b+z)/(\bm{x}_{n}\cdot \bm{x}_{n})$. Finally, we obtain
\begin{align}
\bm{a}=\frac{b+z}{\bm{x}_{n}\cdot \bm{x}_{n}} \bm{x}_{n}.
\end{align}

Although there are several choices for the distribution for $b$, we here use a normal distribution. We summarize the procedure in Algorithm~\ref{alg3}. The output parameters $\{c_j\}$ are determined by the ridge regression.

\begin{algorithm}[H]
\caption{Modified sampling algorithm II (Sampling of $b$)}
\label{alg3}
\begin{algorithmic}[1]
\STATE Choose $n$ according to the probability $\{\omega_{n}\}$.
\STATE Draw $\zeta\sim \mathrm{Beta}(\alpha,\beta)$ and $\gamma\sim \mathrm{Bernoulli}(\gamma;p=0.5)$
\STATE $z\leftarrow(-1)^{\gamma}\zeta$
\STATE Draw $b\sim\mathcal{N}(0,\delta)$
\STATE $r\leftarrow (b+z)/(\bm{x}_{n}\cdot \bm{x}_{n})$
\STATE $\bm{a}\leftarrow r\bm{x}_{n}$
\STATE Return $(\bm{a},b)$
\end{algorithmic}
\end{algorithm}

\subsection{Numerical experiments}

\subsubsection{One dimensional example}
\label{subsubsection_TSC}

As demonstrations, we here perform numerical experiments for the TSC example. The settings are the same as Sect.~\ref{subsec_numerical_example_TSC}. We use $\delta = 15$ both in Algorithm~\ref{alg2} and Algorithm~\ref{alg3}. The output parameters $\{c_j\}$ are determined by the ridge regression with the regularization parameter of $0.01$.

\begin{figure}[hbtp]
\begin{center}
\includegraphics[width=65mm]{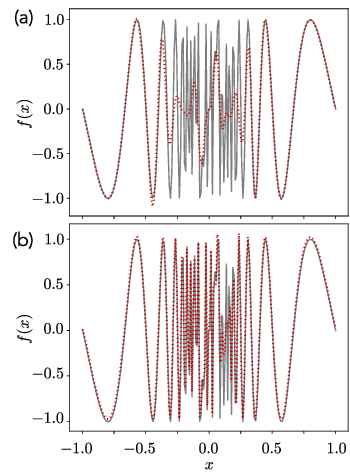}
\caption{(Color online) (a) $f(\bm{x})$ obtained by Algorithm~\ref{alg2} and the ridge regression. (b) $f(\bm{x})$ obtained by Algorithm~\ref{alg3} and the ridge regression. The solid and dotted curves correspond to the original functions and the obtained ones, respectively.}
\label{fig_proposals}
\end{center}
\end{figure}

Figure~\ref{fig_proposals} shows the obtained curves with Algorithms~\ref{alg2} and \ref{alg3}. In particular, Algorithm~\ref{alg3} yields better results near the complicated shape around $x = 0$, and the results indicate that the selection of $\bm{a}$ could be more crucial than $b$. Note that the information from the selected data $\bm{x}_n$ is included to choose the amplitude of $\bm{a}$ in Algorithm~\ref{alg3}, while Algorithm~\ref{alg2} determines it by sampling from the normal distribution. Hence, Algorithm~\ref{alg3} reflects the data characteristics in the choice of the amplitude of $\bm{a}$, which would reduce the arbitrariness of the density function. Although its theoretical analysis will be hopeful in the future, this finding seems valid in some other cases; see a higher dimensional case in Sect.~\ref{subsubsec_MNIST}.

Note that it is possible to see the sampling with the ridgelet transform as an initialization of parameters in neural networks. Hence, we next perform a learning process for neural networks, in which the initial parameters are determined by the above algorithms. In the learning stage, we set the loss function as \texttt{torch.nn.MSELoss}, the optimization method as \texttt{torch.optim.Adam} in \verb|PyTorch| \cite{PyTorch}. The learning rate is set to be $0.001$. For comparison, we also construct a neural network with the same structure but with conventional random initialization. The initial parameters for $\{a_j\}, \{b_j\}$ and $\{c_j\}$ are drawn from $\mathcal{N}(0,1)$; we tried several settings, and this variance gave reasonably good results.

\begin{figure}[tp]
\begin{center}
\includegraphics[width=90mm]{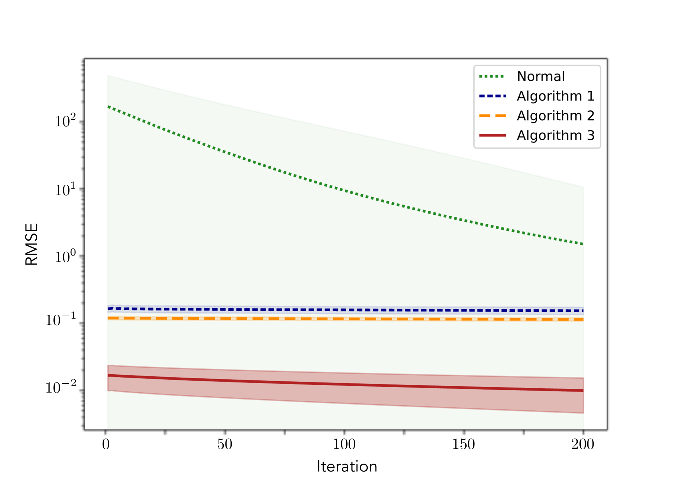}
\caption{(Color online) The training errors in the learning processes for the TSC function. The vertical axis shows root mean squared errors, and the horizontal axis indicates the iteration steps. The error regions are drawn based on the standard deviations.}
\label{fig_result_1d}
\end{center}
\end{figure}

Figure~\ref{fig_result_1d} summarizes the mean training results for $100$ generated neural networks. The vertical axis shows root mean squared errors, and the horizontal axis indicates the iteration steps. The error regions are depicted based on the standard deviations. As expected, Algorithms~\ref{alg1}, \ref{alg2}, and \ref{alg3} significantly reduce the initial errors compared to the conventional random initialization, and they also yield small standard deviations, which suggests the sampling algorithms work well.

\subsubsection{Classification for high dimentional inputs}
\label{subsubsec_MNIST}

Next, we  provide numerical results for high-dimensional input cases; as in the previous study \cite{Sonoda2014}, we employ the MNIST dataset consisting of $60,000$ training examples and $10,000$ test examples \cite{LeCun1998,Deng2012}. Each input is a grayscale image of a handwritten digit of $784$ pixels. The corresponding output is one of the ten numbers from 0 to 9. As in Ref.~\cite{Sonoda2014}, each number is represented as a 10-dimensional vector; the components of vectors are randomly selected with probability equal to 1 and 0. The regressions for Algorithms~\ref{alg1}, \ref{alg2}, and \ref{alg3} are performed with the label vectors. After initializing the neural network parameters, we add finalization procedures to the neural network; the neural network outputs are standardized and applied to a sigmoid function to obtain the final output vector. The digit of the closest label vector is finally chosen. The number of hidden nodes is $300$. We set the loss function as \texttt{torch.nn.CrossEntropyLoss} and an optimization method as \texttt{torch.optim.Adam}; the learning rate is $0.001$. The settings for Algorithms~\ref{alg1}, \ref{alg2}, and \ref{alg3} are the same as those in Sect.~\ref{subsubsection_TSC}, except that a simple linear regression is used for Algorithm~\ref{alg1} as in Ref.~\cite{Sonoda2014}.

\begin{figure}[t]
\begin{center}
\includegraphics[width=90mm]{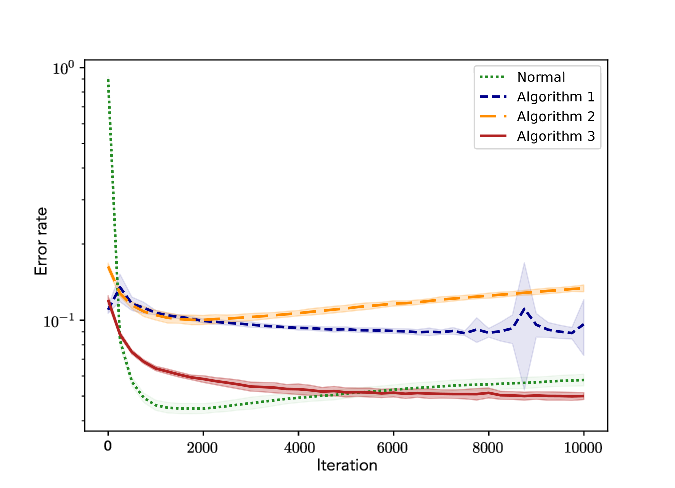}
\caption{(Color online) The classification error rates in the learning processes for the MNIST test dataset. The vertical axis shows root mean squared errors, and the horizontal axis indicates the iteration steps. The error regions are drawn based on the standard deviations.}
\label{fig_result_mnist}
\end{center}
\end{figure}

Figure~\ref{fig_result_mnist} depicts the classification errors for the test examples; means and standard deviations of 10 generated neural networks are shown. As similar to the TSC cases in Sect.~\ref{subsubsection_TSC}, Algorithms~\ref{alg1}, \ref{alg2}, and \ref{alg3} significantly reduce the initial errors compared to the conventional random initialization. Note that the conventional parameter initialization with the normal distributions yields the best accuracy around 1000 iterations steps, although it shows overlearning behavior. We performed several other parameter fittings, and the tendency is not so changed. It is difficult to explain the behaviors for the results with Algorithms~\ref{alg1}, \ref{alg2}, and \ref{alg3} theoretically, but Algorithm~\ref{alg3} yields a reasonable result even in this high-dimensional case. Hence, the sampling for the magnitude of $\bm{a}$ could be more crucial than $b$, as conjectured in Sect.~\ref{subsubsection_TSC}.

\section{Conclusion}
\label{sec_conclusion}

We provided a new perspective on the ridgelet transform from the aspect of the importance sampling. The perspective enables us to avoid the linear regression procedure in the previous study, which recovers rough function shapes. We also confirmed that the proposed two approximated algorithms work well, taking advantage of the arbitrariness of the sampling distribution. In particular, it was beneficial to use the information of datasets in the sampling step of $\bm{a}$. This numerical results indicate that the sampling of parameter $\bm{a}$ could be more crucial than $b$.

There are some remaining works in the future. For example, it is crucial to provide theoretical discussions on the importance of $\bm{a}$ compared to $b$. It is also essential to discuss the choice of $\rho^{(2)}(\bm{a},b)$ in Eq.~\eqref{eq_IS_2}; if we can determine $\rho^{(2)}(\bm{a},b)$ from the datasets, we could avoid the regression procedure for the output parameters. In addition, there are discussions on the ridgelet transform and deep neural networks \cite{Sonoda2019,Sonoda2023}. As far as we know, there are no practical sampling algorithms based on the ridgelet transform for the deep cases. In the future, we expect to clarify connections between the parameter distributions and integral representations of functions from a physics viewpoint.

\begin{acknowledgments}

This work was supported by JSPS KAKENHI Grant Number JP21K12045.
\end{acknowledgments}


\begin{thebibliography}{30}

\bibitem{Barron1993}
A.R. Barron, IEEE Trans. Info. Theory \textbf{39}, 930 (1993).
  
\bibitem{Murata1996}
N. Murata, Neural Networks \textbf{9}, 947 (1996).


\bibitem{Sonoda2014}
S. Sonoda and N. Murata, Lecture Notes in Computer Science \textbf{8681}, 539 (2014).


\bibitem{Denoeux1993}
T. Denoeux and R. Lengell{\'e}, Neural Networks \textbf{6}, 351 (1993).

\bibitem{Freitas2000}
J.F.G. de Freitas, M. Niranjan, A.H. Gee, and A. Doucet, Neural Comput. \textbf{12}, 955 (2000).


\bibitem{LeCun2012}
Y. LeCun, L. Bottou, G.B. Orr, and K.-R. M{\"u}ller, Lecture Notes in Computer Science \textbf{7700}, 9 (2012).


\bibitem{Huang2004}
G.-B. Huang, Q.-Y. Zhu, and C.-K. Siew, Proc. 2004 IEEE Int. Joint Conf. Neural Networks, 2004, p. 985.


\bibitem{Huang2006}
G.-B. Huang, Q.-Y. Zhu, and C.-K. Siew, Neurocomputing \textbf{70}, 489 (2006).

\bibitem{Wang2022}
J. Wang, S. Lu, S.-H. Wang, and Y.-D. Zhang, Multimed. Tools Appl. \textbf{81}, 41611 (2022).

\bibitem{Robert_book}
C. P. Robert and G. Casella, Monte Carlo Statistical Methods, 2nd ed (Springer, New York, 2004).

\bibitem{Sonoda2017} 
S. Sonoda and N. Murata, Applied and Computational Harmonic Analysis \textbf{43}, 233 (2017).


\bibitem{PyTorch}
\url{https://pytorch.org} (accessed on June 24, 2024).

\bibitem{LeCun1998}
Y.  LeCun, L. Bottou, Y. Bengio, and P. Haffner, Proc. IEEE \textbf{86}, 2278 (1998).
\bibitem{Deng2012}
L. Deng, IEEE Signal Process. Mag. \textbf{29}, 141 (2012).


\bibitem{Sonoda2019}
S. Sonoda and N. Murata, J. Machine Learn. Res. \textbf{20}, 1 (2019).

\bibitem{Sonoda2023}
S. Sonoda, Y. Hashimoto, I. Ishikawa, and M. Ikeda, arxiv:2310.03529.


\end{thebibliography}
\end{document}